\documentclass{WileyMSP-template}

\usepackage{xcolor}
\usepackage[acronym]{glossaries}
\usepackage{makecell}

\newacronym{pcs}{PCs}{Pacinian corpuscles}
\newacronym{xr}{XR}{extended reality}
\newacronym{fps}{FPS}{frames per second}
\newacronym{api}{API}{application programming interface}
\newacronym{udp}{UDP}{user datagram protocol}
\newacronym{hmd}{HMD}{head-mounted display}
\newacronym{vla}{VLA}{vision-language-action}

\begin{document}


\title{High-Bandwidth Biomimetic Finger for Tactile-Transparent Remote Texture Sensing}

\maketitle


\author{Shuang Yang$^1$}
\author{Fuyuan Liu$^1$}
\author{Yitian Shao$^{1,2}$}


\dedication{}

\begin{affiliations}
$^1$School of Computer Science and Technology, Harbin Institute of Technology, Shenzhen, Shenzhen 518055, China.\\
$^2$State Key Laboratory of Smart Farm Technologies and Systems, Harbin 150006, China.\\
Correspondence: Fuyuan Liu (liufy@hit.edu.cn) and Yitian Shao (shaoyitian@hit.edu.cn)
\end{affiliations}



\begin{abstract}
High-fidelity tactile feedback is essential for robotic teleoperation, enabling precise manipulation and critical decision-making. While biomimetic fingertip sensors can capture surface-texture features, how their design shapes the tactile transparency of rendered feedback remains poorly understood. This paper presents a biomimetic fingertip replicating the human finger's multilayer mechanical gradient and fingerprint morphology, with an embedded high-sensitivity inertial measurement unit capturing texture-induced vibrations for remote vibrotactile rendering. Three variants are compared with the human fingertip through temporal and spectral analyses and a user study spanning three perceptual dimensions, establishing a transmission chain from fingertip design through signal characteristics to tactile transparency. The multilayer mechanical gradient yields the clearest improvements in signal intensity and texture-feature representation, whereas a stiffer skin layer amplifies vibration intensity at the expense of feature representation. The perceptual dimensions draw on distinct signal attributes: roughness on energy scaling, granularity on spectral shape, and repetitivity on characteristic-peak representation. These findings offer design principles for application-specific fingertips in robotic teleoperation.
\end{abstract}

\section{Introduction}
Humanoid robots have advanced rapidly over the past few years, yet their tactile sensing capabilities still lag far behind vision. Current robotic tactile sensors differ fundamentally from the human hand; if a human operator attempts to perceive the signals collected by a robot finger, it results in a mismatched sensation. Thus, robot hands need a better tactile sensing mechanism that offers high tactile transparency, allowing a teleoperator to naturally feel exactly what the robot touches. This crucial area has yet to be fully explored.

To achieve tactile transparency, it is essential to understand how physical stimuli map to human perception. 
Neurophysiological studies show that population responses evoked by different textures exhibit characteristic high-frequency-to-low-frequency energy ratios, establishing a spectral-balance coding system. Discrimination of vibration stimuli depends primarily on the energy balance between these low- and high-frequency components \cite{Bernard_2024}. 
From an engineering standpoint, the spectral characteristics of vibrations have been confirmed as the primary cues enabling discrimination between real textures and those rendered by tactile devices, underscoring the central role of vibration spectra as carriers of tactile information \cite{Fagiani_2012,Felicetti_2023}. Taken together, whether a tactile sensor can reproduce human perceptual realism hinges on its ability to capture spectral energy distributions similar to those of the human fingertip across a broad frequency band.
Therefore, existing designs utilize the structural interlocking of a rigid epidermis and a flexible dermis at the dermal papillae, forming a mechanical gradient from the surface to deeper layers \cite{Choi_2019,Cei_2025}; periodically arranged fingerprint ridges enhance frictional grip and spatial recognition \cite{Jarocka_2021,Hao_2024}, and convert surface textures into frequency-specific vibrations \cite{Zhao_2021}. 

Studies of fingertip contact mechanics and finite-element analysis have shown that the natural vibration modal frequencies of fingertip tissue lie in the range of 100-260 Hz, with the fourth-order mode ($\sim$225 Hz) responding most prominently to normal excitation \cite{Serhat_2021,Serhat_2024}. 
The upper frequency limit of most flexible sensors, however, is only 100-200 Hz \cite{Zhang_2022}; they therefore lack the high-frequency components that serve as the principal information carriers in spectral-balance coding. Current biomimetic sensors predominantly mimic the multi-layer structure and tune modulus gradients to maximize texture discrimination accuracy \cite{Zhao_2021,Dai_2022,Bai_2023,Qin_2024}. Under this paradigm, spatial resolution and static texture classification accuracy have been approached or surpassed the level of the human finger \cite{Ouyang_2024,Kang_2024}. 
A fundamental question, however, remains largely unexplored: how do the structural configuration, geometric morphology, and material properties of a biomimetic fingertip influence the spectral characteristics of the captured vibrotactile signals, and thereby affect the tactile transparency of vibrotactile feedback.

In this work, we design and compare three biomimetic fingertip prototypes with heterogeneous structures, integrating a high-sensitivity inertial measurement unit (IMU) within a three-layer structure that mimics the mechanical gradient and mechanoreceptors of the human finger. We fabricated the fingertips using 3D printing and a low-cost injection-molding process, and evaluated them by sliding across 11 textured surfaces at controlled speeds. The vibrotactile signals captured by each biomimetic fingertip were compared with those of the human fingertip across temporal features and spectral features. 
Moreover, evaluating vibrotactile feedback requires looking beyond physical metrics to encompass perceptual dimensions. Prior studies demonstrate that even if a synthesized tactile signal structurally differs from a real-world finger vibration and lacks its exact spectral details, it can still evoke a highly realistic sensation \cite{rosenkranz2023perceptual}.
Therefore, we evaluate the perceptual aspect of the captured signals through user study aiming to allow a user to feel the touch of robot finger, as shown in Figure 1(a). Our studies reveal how structural configuration, material stiffness, and fingerprint microstructure affect tactile transparency, offering design principles for biomimetic tactile sensors that aim to deliver perceptually faithful feedback in robotic teleoperation.

\section{Related Work}

\subsection{Structural Design of Biomimetic Finger}

Inspired by the human fingertip, biomimetic tactile sensors have systematically emulated the morphological features of the human finger from two aspects, geometric structure and material selection \cite{Andrussow_2023}, which formed a comprehensive trajectory from morphological biomimicry and material innovation to system-level integration \cite{Tanaka_2019,Geng_2023}.

For macroscopic geometry, sensor shape has evolved from simple flat plates to curved, finger-shaped structures that more faithfully reproduce fingertip-to-surface contact geometry \cite{Andrussow_2023}. In terms of microscopic surface texture, fingerprint patterns exhibit a diversifying trend: parallel straight ridges \cite{Dai_2022}, sine-wave and concentric-circle ridges \cite{Kim_2020}, whorl- and loop-type textures \cite{Hao_2022}. Non-fingerprint patterns such as honeycomb and linear textures have also been systematically compared for different application scenarios, aiming to identify the optimal texture geometry for vibration signal modulation \cite{Guo_2024}. The fundamental function of these textures is to exploit periodic contact-separation cycles between ridge protrusions and the surface, thereby modulating spatial texture information into time-varying mechanical signals.

Multi-layer structural biomimicry lies at the core of morphological imitation. To reproduce the skin-subcutaneous tissue-bone mechanical gradient of the human fingertip, the typical configuration employs a hard epidermal layer (SU-8, PI, elastic modulus on the order of GPa), a soft intermediate layer (PDMS, Ecoflex, on the order of MPa), and a rigid base layer \cite{Choi_2019,Tanaka_2019,Navaraj_2019,Hao_2025}. In this configuration, the hard fingerprint ridges undergo negligible deformation upon surface contact and transmit surface profile information to the subcutaneous tissue; the soft intermediate layer permits local tilting and elastic recovery of the ridge structures under shear forces, converting tangential friction into normal vibration \cite{Choi_2019}. Although the efficacy of these structural configurations in amplifying texture-induced signals is well established, their role in modulating the spectral properties of transduced vibrations, and consequently the perceptual fidelity of tactile feedback, is yet to be explored.

\subsection{Sensing Mechanisms of Biomimetic Finger}

The tactile perception of the human finger depends on four types of mechanoreceptors, which form two complementary coding channels. SA-I (Merkel) and SA-II (Ruffini) encode near static pressure and skin stretch; FA-I (Meissner, $\sim$\,5 -- 50\,Hz) and FA-II (Pacinian, around $\sim$\,40 -- 400\,Hz) encode low-frequency slip and high-frequency vibration \cite{johansson_2009}. The mechanical design of full-bandwidth tactile sensing for biomimetic fingers is, in essence, a search for transduction schemes that correspond to these functional channels.

For the spatial coding, piezoresistive, capacitive, and optical sensing are the most widely adopted approaches. 
Integrated within biomimetic fingertip substrates, these sensors capture contact pressure distribution and normal force during object contact, enabling static texture shape recognition and precision grip force control \cite{Ouyang_2024, Navaraj_2019}. 
Magnetosensitive and inductive schemes further extend the sensing scope to multi-axis force and torque, which are critical for detecting slip and characterizing fine contact mechanics \cite{Wu_2018, Xu_2026}. 
Collectively, these approaches have driven spatial resolution and static contact pattern recognition to levels approaching or surpassing the human finger \cite{Ouyang_2024}.

For the temporal coding, piezoelectric sensing has emerged as the predominant modality due to its rapid response and high bandwidth \cite{Zhang_2022, Qiu_2020, Kim_2020, Choi_2019}. Alternative mechanisms, such as triboelectric \cite{Guo_2024, Qiao_2023} and iontronic \cite{Bai_2023} schemes, have also been investigated to provide complementary advantages, including self-powered operation and single-sensor dual-channel detection, respectively. To capture these rapid dynamics, the sampling rates of FA-mimicking sensors typically range from 1-25 kHz \cite{Kim_2020, Qin_2024, Shi_2023, Qin_2023, Kovenburg_2023, Tanaka_2019}. In practice, a low-pass filter with a cutoff frequency around 50 Hz is applied during signal conditioning to suppress high-frequency noise and power-line interference \cite{Kim_2020, Qin_2024}. Since texture-induced vibrations encode multi-scale surface features across a broad frequency range rather than in a single dominant frequency \cite{Kim_2020}, capturing this spectral information requires high bandwidth. Accordingly, several studies have extended the effective bandwidth to approximately 250 Hz \cite{Rostamian_2022, Bai_2023, Sankar_2025}, 600 Hz \cite{Zhang_2022, Qiu_2020}, and up to 1000 Hz \cite{Qiu_2024_sciadv, Tanaka_2019}. Frequency-domain features, including FFT spectra, wavelet coefficients, and spectral centroid, have thus been widely adopted as primary inputs for texture classification \cite{Bai_2023, Shi_2023, Qin_2023, Qin_2024, Kovenburg_2023}. Wideband signal capture has further enabled the detection of high-frequency stimuli \cite{Zhang_2022} and the deconvolution of complex loading processes \cite{Qiu_2020}. 

While signal acquisition techniques covering the full bandwidth of human tactile sensing have advanced significantly, the degree to which robotic tactile signals mirror human perception remains underexplored. Existing research primarily targets spectral feature reproduction for texture discrimination, offering little insight into the perceptual realism of the acquired signals.

\subsection{Evaluation Metrics of Biomimetic Finger} 

With biomimetic form providing the physical foundation and sensing mechanisms defining the signal transduction pathways, existing work has concentrated on amplifying texture-induced vibration signals and improving texture discrimination capability. Three progressive levels of performance advancement can be identified.

The first level is the enhancement of signal capture through structural optimization. The basic strategy is to combine hard epidermal ridges (GPa-scale) with a soft substrate (MPa-scale), exploiting the stiffness mismatch to convert tangential friction into normal vibration. Studies have shown that this ``hard-ridge/soft-substrate'' design amplifies vibration signal amplitude by nearly an order of magnitude (6.85$\times$), enabling the sensor to resolve texture periods on the order of hundreds of microns and height differences on the order of tens of microns \cite{Choi_2019}. Fingerprint ridges of different geometries amplify vibration features in complementary frequency bands, and hybrid patterns provide richer texture discrimination information than single patterns \cite{Kim_2020}.

The second level is the pursuit of ultra-high spatial resolution once vibration signals have been effectively captured. Fingerprint-inspired electronic skin based on single-electrode triboelectric nanogenerators exploits the sub-micron sensitivity of interfacial charge transfer to contact-area variations, pushing the minimum resolvable texture feature size to 6.5 $\mu$m \cite{Zhao_2021}. Iontronic slip sensors based on the electric double-layer mechanism, combined with hierarchically microstructured ionogels, achieve recognition rates of 100\% and 98.9\% for 20 types of commercial textiles under fixed and random sliding speeds, respectively \cite{Bai_2023}.

The third level is the surpassing of human finger performance on specific metrics. A high-density piezoresistive sensor array (32 $\times$ 32 pixels) achieved a response time of approximately 7 ms and a two-point discrimination threshold of 1.5 mm (versus $\sim$2-3 mm for the human finger), with a contact pattern recognition accuracy (99.0\%) far exceeding that of human participants (69.1\%) \cite{Ouyang_2024,Lee_2024}. Capacitive electronic skin replicating the real fingerprint topologies of historical figures provided, for the first time, experimental evidence that different fingerprint morphologies produce significantly different vibration conduction characteristics, underscoring the critical role of fingerprint geometry as a tactile signal modulation interface \cite{Hou_2024}.

The three levels described above reflect a mature evaluation paradigm centered on texture discrimination accuracy. However, prior research suggests that improperly incorporating fingerprint information into vibrotactile signals may degrade perceptual authenticity \cite{Weiland_2024}. Consequently, we argue that texture discrimination performance cannot serve as the sole objective criterion for evaluating sensing performance. Instead, we aim to investigate the tactile transparency of biomimetic sensors by accounting for temporal waveforms, spectral features, and human perceptual evaluation.

\begin{figure}[!t]
  \includegraphics[width=16cm]{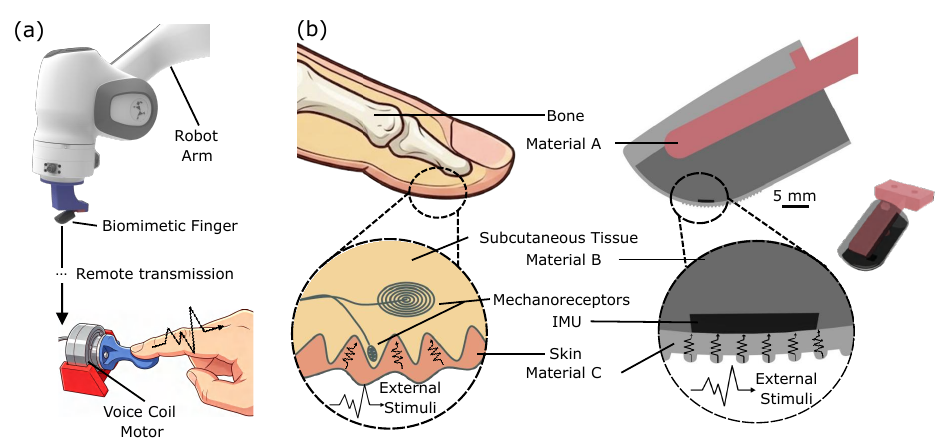}
  \caption{A biomimetic fingertip capable of capturing micron-level textural details to deliver high-fidelity vibrotactile feedback. (a) Vibrotactile signals acquired by the biomimetic finger mounted on a Franka Research 3 robotic manipulator are remotely transmitted to the operator via a voice coil motor. (b) Structural comparison between the human fingertip anatomy and the proposed biomimetic design. }
  \label{fig1}
\end{figure}

\section{The Biomimetic Fingertip}
\subsection{Mechanical Design}
Telerobotics enables remote physical interaction across spatial boundaries, where high-fidelity tactile feedback is essential to overcome the dexterity limitations of conventional visual feedback during delicate manipulation. Aiming to provide informative and high-fidelity tactile feedback to human teleoperators, we designed a biomimetic fingertip that captures vibrotactile signals during surface sliding via an inertial measurement unit (IMU). 

We adopted a biomimetic framework to engineer the mechanical structure of the finger.
The human fingertip comprises five tissue layers with a pronounced mechanical gradient: stratum corneum ($\sim$1.0 MPa), epidermis ($\sim$0.14 MPa), dermis ($\sim$0.08 MPa), hypodermis ($\sim$0.034 MPa), and bone ($\sim$17 GPa) \cite{somer_2015, Serhat_2021}. The stratum corneum and epidermis, spanning a similar stiffness range, are functionally merged into a single skin layer; the dermis and hypodermis are likewise combined as subcutaneous tissue. Accordingly, we adopted a three-layer configuration consisting of a skin layer (0.5 mm thick), a subcutaneous tissue layer, and a rigid bone scaffold, with the IMU (LSM6DSR, STMicroelectronics) embedded centrally between the skin and subcutaneous layers.

To replicate this mechanical gradient, we selected four materials with graded stiffness: Polyethylene Terephthalate Glycol-modified (PETG, $\sim$1.5 GPa) for the rigid bone scaffold; Polydimethylsiloxane (PDMS, $\sim$2.0 MPa) or Thermoplastic Polyurethane-95A (TPU-95A, $\sim$10.0 MPa) for the skin layer; and PDMS or Ecoflex-30 ($\sim$0.1 MPa) for the subcutaneous tissue layer. The outer surface of the skin layer is contoured to replicate the natural curvature of the human fingertip and is patterned with biomimetic fingerprint ridges extracted from optical coherence tomography (OCT) scans. By combining different material assignments across layers, we produced three biomimetic fingertip variants, as detailed in the following subsection.

\begin{figure}[!t]
  \includegraphics[width=\linewidth]{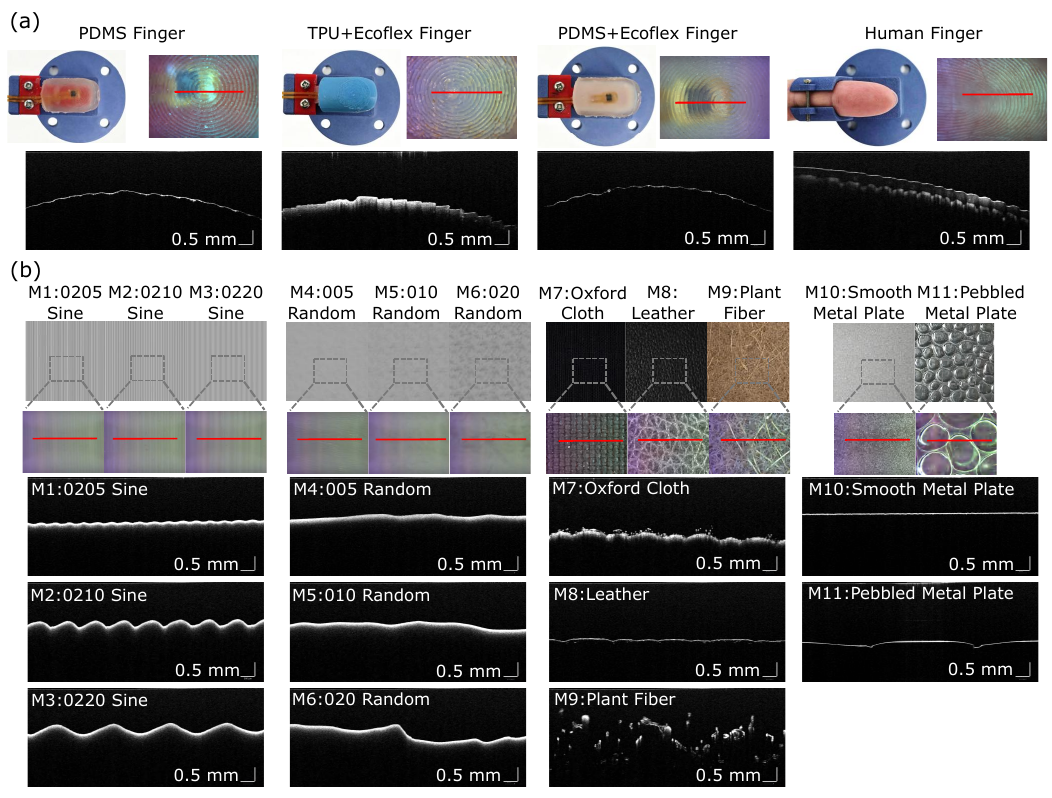}
  \caption{(a) Morphological comparison of three biomimetic fingertips with the human fingertip. Top: Optical macrographs showcasing finger profiles and epidermal ridge patterns; Bottom: Cross-sectional optical coherence tomography (OCT) images of the fingerprints along the sliding direction. (b) Surface characterization of eleven textured substrates. Top: Photographs displaying material surfaces and distinct textures; Bottom: Cross-sectional OCT profiles of the textures captured along the red line indicating the sliding path.}
  \label{fig2}
\end{figure}

\subsection{Fingertip and Textured Surfaces}

Following the layer-material assignments described above,  we fabricated three variants: (i) the single-material PDMS design, (ii) TPU-95A skin with Ecoflex-30 subcutaneous layer, and (iii) PDMS skin with Ecoflex-30 subcutaneous layer, as depicted in Figure 2(a). All three share an identical PETG bone scaffold and geometric model; minor differences in fingerprint ridge contours arise from fabrication process variations, as discussed in Section 3.3.

Figure 2(a) shows cross-sectional OCT images of each fingertip alongside a human fingertip. The PDMS and PDMS+Ecoflex fingertips, both fabricated via molding, exhibit ridge morphologies and overall curvature comparable to those of the human fingertip. The TPU-95A fingertip, fabricated via fused deposition modeling (FDM) 3D printing, shows lower ridge fidelity: the FDM process involves an inherent trade-off between preserving fine ridge geometry and minimizing visible layer-line artifacts, resulting in jagged, stair-stepped ridge contours rather than the designed smooth arc-shaped profile.

Figure 2(b) shows the 11 test surfaces. Six are numerical surfaces fabricated via Polylactic Acid (PLA) 3D printing: three periodic (M1-3, sinusoidal profile, 0.2 mm amplitude, spatial periods of 0.5, 1.0, and 2.0 mm) and three random (M4-6, Gaussian height distribution, RMS heights of 0.05, 0.10, and 0.20 mm). The remaining five are natural surfaces selected to span a broad range of surface morphologies encountered in everyday objects: Oxford cloth (M7, woven textile with regular periodic structure), leather (M8, natural grain with moderate roughness), plant fiber (M9, highly disordered fibrous surface with large roughness), smooth metal plate (M10, polished, near-featureless), and pebbled metal plate (M11, regularly spaced convex protrusions).

\begin{figure}[!t]
  \includegraphics[width=\linewidth]{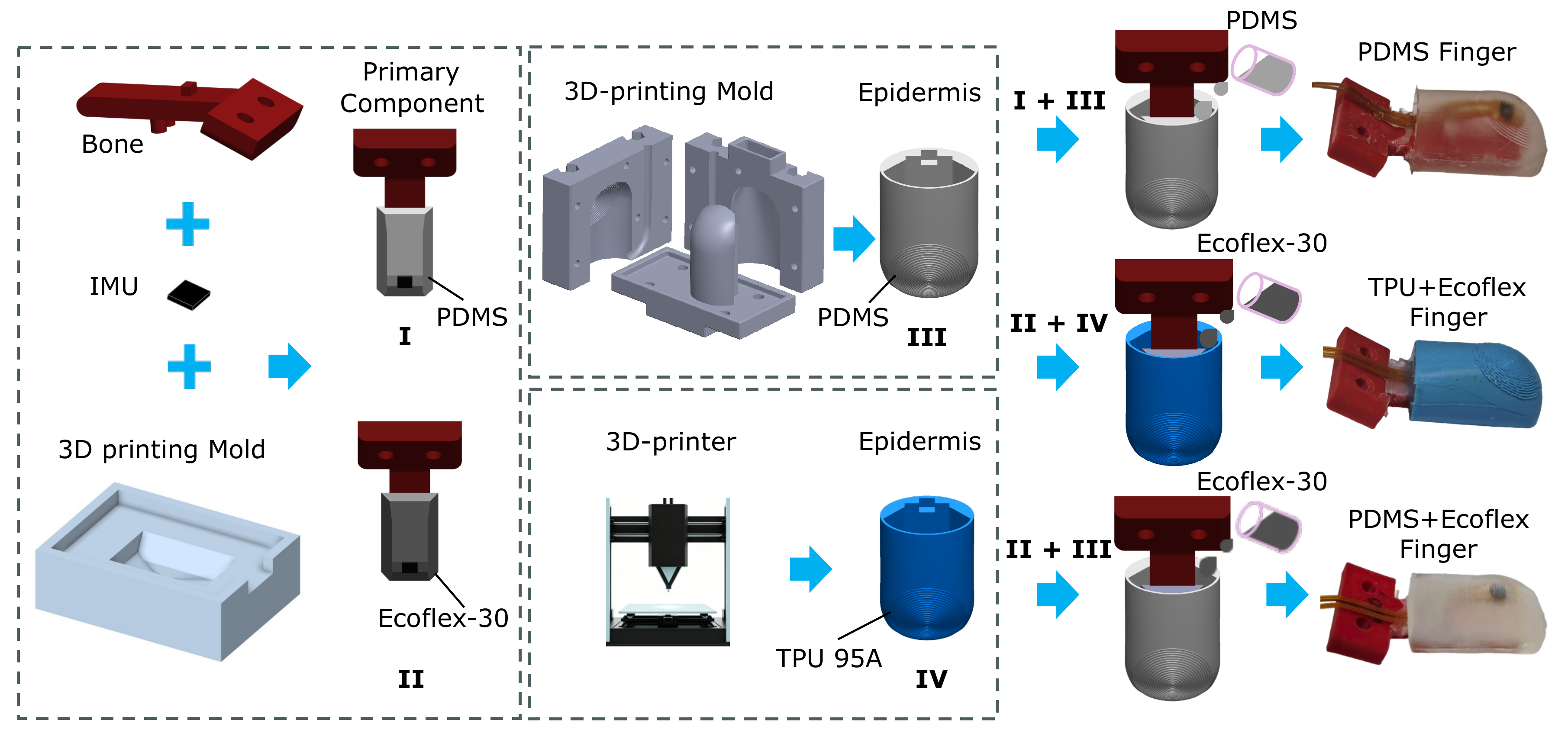}
  \caption{Schematic illustration of the fabrication process for biomimetic fingers with distinct designs. I)/II) Modeling of key components to integrate the 3D-printed bone and IMU unit, using PDMS or Ecoflex-30 as the soft filling material. III) Fabrication of PDMS skin via molding with custom 3D-printed molds. IV) Fabrication of TPU-95A skin via direct 3D printing.
  }
  \label{fig3}
\end{figure}

\subsection{Fabrication}

To fabricate the tactile sensors, we designed custom molds for injection molding of the multilayer elastomer structures. As shown in Figure 3, the fabrication procedure consists of three main steps: 1) The bone structure and IMU are assembled together via molding with PDMS or Ecoflex-30 inside the PLA mold to fix the relative position between the IMU and bone structure; 2) The epidermal layer is manufactured either by injection molding PDMS using the PLA mold or by 3D printing TPU-95A; 3) The assembly obtained from Step 1 and the epidermal layer from Step 2 are bonded together via molding with the same elastomer material used in Step 1. By adopting different material combinations for the base assembly and epidermal layer, we finally obtained three types of biomimetic fingers.

It is worth noting that both the bone structure and PLA molds are fabricated via 3D printing (H2D, Bamboo). The Ecoflex-30 mixture was prepared by combining the two components at a 1:1 volume ratio, degassed under vacuum, and then cured at room temperature for 1 hour after molding. For the PDMS mixture, the base and curing agent are blended at a 10:1 ratio and degassed under vacuum. The mixture is placed in a vacuum drying oven and heated at 60\textsuperscript{◦}C for 1 hour to complete solidification.

\subsection{Data Preprocessing}

The data from biomimetic and human finger were acquired using a computer (Raspberry Pi 5, the Raspberry Pi Foundation) and transferred to the host computer via WiFi. As a common preprocessing step, the data were demeaned, and low-pass filtered using an 8th-order Butterworth filter with a cutoff frequency of 300 Hz at a sampling rate of 7000 Hz; this suppresses high-frequency noise and confines the signal to the band relevant to human vibrotactile perception.

Three analysis-specific normalization strategies were then applied according to the downstream purpose. For similarity analyses, including the waveform cross-correlation (Section 4.2) and the spectral cosine similarity (Section 4.4, Figure 6), signals were normalized to mitigate the influence of amplitude scale on similarity assessment. For signal visualization, including the waveform display (Section 4.2, Figure 4(c)) and the spectral plots (Section 4.3, Figure 5), signals were used without further normalization to preserve amplitude differences across fingertips. For vibrotactile feedback rendering in the user study (Section 5), signals were aligned to a common group RMS level, divided by 5 times this RMS value, and clipped to [-1, +1] to balance amplitudes across surfaces and speeds while matching the input dynamic range of the motor. This preprocessing pipeline constitutes the methodological foundation of this work, and all subsequent analyses were performed on the preprocessed dataset according to the analysis-specific normalization described above.


\section{Evaluation Experiment}
To evaluate how the design features of different biomimetic fingertips influence tactile transparency, we conducted sliding experiments with fingertips mounted on a robot arm platform (FR3, Franka); a high-sensitivity IMU (LSM6DSR, STMicroelectronics) records texture-induced vibrations. We compared the vibrotactile signals captured by each biomimetic fingertip and the human fingertip in both the temporal and spectral domains, identifying design-dependent differences in signal fidelity. These analyses also informed the design of the subsequent user study.

\subsection{Data Acquisition with Fingers}
Experiments were conducted on a robotic arm, shown in Figure 4(a). A customized 3D-printed end-effector held both the biomimetic fingertip and the human subject's index finger, translating them at constant velocity across each test surface. The IMU embedded within the biomimetic fingertip (Section 3.1) recorded vibrotactile signals during sliding. An IMU of the same model was affixed to the subject's index finger, serving as the reference for human tactile sensing.

\begin{figure}[!t]
  \includegraphics[width=\linewidth]{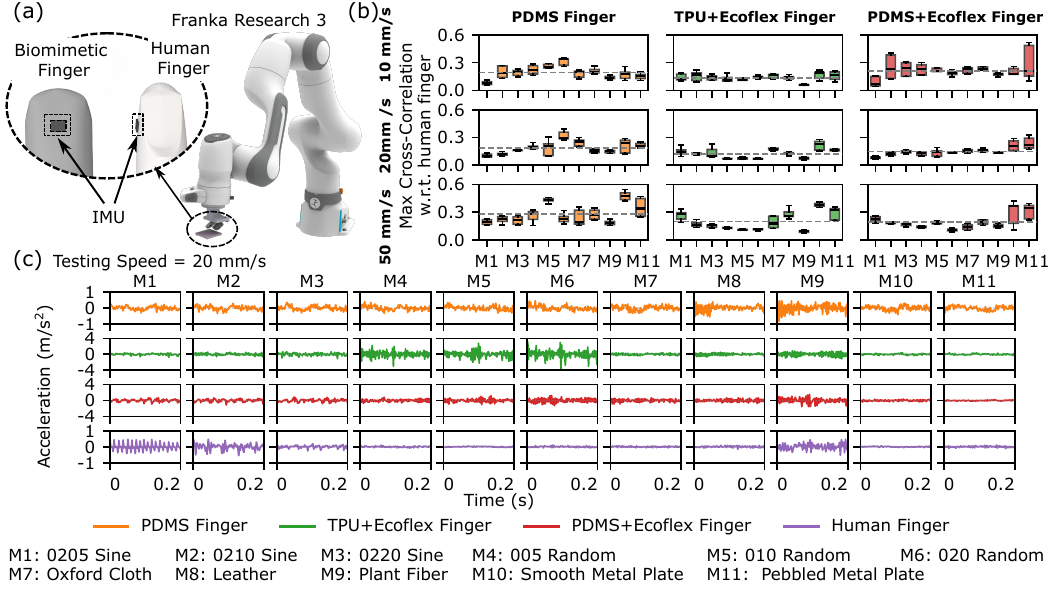}
  \caption{Experimental setup and signal comparison overview. (a) The Franka Research 3 robotic arm with a biomimetic fingertip mounted on the customized end-effector; an IMU embedded within the biomimetic fingertip and a second IMU affixed to the human subject's index finger recorded vibrotactile signals during sliding. (b) Maximum cross-correlation coefficients between the vibrotactile signals of each biomimetic fingertip and the human fingertip, computed across 11 textured surfaces at three sliding speeds (10, 20, and 50 mm/s); the grey dashed lines indicate the mean correlation coefficient across all 11 surfaces. (c) Representative temporal waveforms at 20 mm/s for the three biomimetic fingertips and the human fingertip.}
  \label{fig4}
\end{figure}

Under robotic arm control, each finger was pressed against the test surface to a depth of 0.5 mm and translated forward 40 mm at three constant speeds: 10, 20, and 50 mm/s. This procedure was repeated three times per speed, across 11 textured surfaces and 4 fingers (three biomimetic variants plus the human finger), yielding 396 trials in total. Prior to each trial involving the human finger, the subject wiped the fingerpad with alcohol and allowed it to dry for 5 minutes; the finger was then inserted through the loop of the end-effector and secured by tightening the bolt, as shown in Figure 2(a).

\subsection{Temporal Analysis of Vibrotactile Signals}

To evaluate tactile transparency at the signal level, we adopted the maximum cross-correlation coefficient as a quantitative similarity metric between biomimetic and human fingertip signals. Following the preprocessing described in Section 3.4, signals were normalized to mitigate the influence of amplitude scale on waveform similarity assessment. For each pair of biomimetic and human recordings, the maximum cross-correlation coefficient was computed by searching over a bounded range of time lags. For each (finger, surface, speed) condition, 9 similarity values were obtained from the 3 $\times$ 3 pairwise combinations of biomimetic and human repetitions, forming one box in Figure 4(b). The figure comprises nine subplots (3 fingers $\times$ 3 speeds), each displaying 11 boxes corresponding to the 11 tested surfaces and representing 99 correlation values in total; the grey dashed line in each subplot indicates the mean of these 99 values.

As indicated by the overall means (Table 1; grey dashed lines in Figure 4(b)), the PDMS fingertip exhibits relatively high overall similarity across all three speeds, outperforming the others at 20 and 50 mm/s (0.189 and 0.281). The PDMS+Ecoflex fingertip also demonstrates competitive overall results, achieving the highest rank at 10 mm/s (0.210), whereas the TPU+Ecoflex variant yields the lowest means at 10 and 20 mm/s (0.137 and 0.125). Furthermore, three distinct surface-specific patterns emerge: the PDMS fingertip maintains superior similarity on random surfaces (M4-6); the PDMS+Ecoflex fingertip excels on periodic surfaces (M1-3) at 10 mm/s; and it also dominates on the two metal surfaces (M10, M11) regardless of the sliding speed.

\begin{table}[!htbp]
  \caption{Mean maximum cross-correlation coefficients between the vibrotactile signals of each biomimetic fingertip and the human fingertip at three sliding speeds. Each value averages 99 coefficients (11 surfaces $\times$ 3 $\times$ 3 repetition pairs), corresponding to the grey dashed lines in Figure 4(b).}
  \label{tab1}
  \centering
  \begin{tabular}{@{}lccc@{}}
    \hline
    Fingertip & 10 mm/s & 20 mm/s & 50 mm/s \\
    \hline
    PDMS & 0.193 & 0.189 & 0.281 \\
    PDMS+Ecoflex & 0.210 & 0.149 & 0.193 \\
    TPU+Ecoflex & 0.137 & 0.125 & 0.200 \\
    \hline
  \end{tabular}
\end{table}

Waveform analysis was conducted on a 0.2 s segment (0.6 to 0.8 s after sliding onset), extracted from the steady-contact phase within the middle of the 40 mm trajectory where both contact force and velocity had stabilized. Among the four tested fingertips, the single-material PDMS design yielded the weakest signals. In contrast, the TPU+Ecoflex and PDMS+Ecoflex fingertips, which feature multilayer structures with mechanical gradients, generated more intense vibrations. This observation aligns with prior findings on vibration amplification in hard-ridge/soft-substrate configurations \cite{Choi_2019}. Surface-specific analyses further revealed distinct features: on random surfaces, the TPU+Ecoflex design exhibited a stronger signal response than the PDMS fingertip; on periodic surfaces, both the PDMS+Ecoflex and human fingertips displayed pronounced periodic waveform signatures. Furthermore, the signal energy of the PDMS+Ecoflex fingertip followed surface roughness variations across both random and natural (M9-11) surfaces.

\subsection{Spectral Features Analysis}
\begin{figure}[!t]
  \includegraphics[width=\linewidth]{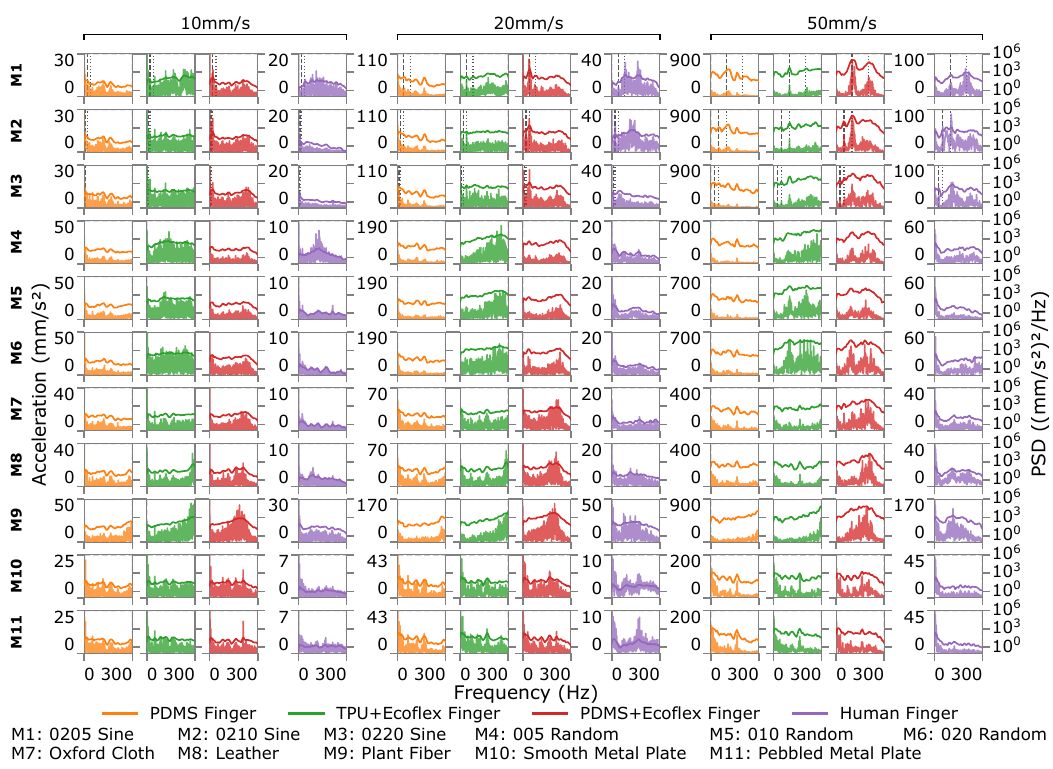}
  \caption{Spectral analysis of vibrotactile signals captured by the three biomimetic fingertips and the human fingertip across 11 textured surfaces (M1-11, one per row), indicated by line colors. The semi-transparent and solid lines represent the FFT spectrum and the Welch power spectral density displayed on a logarithmic axis, respectively. Three blocks of columns correspond to sliding speeds of 10, 20, and 50 mm/s. The dashed lines in subplots of periodic surfaces (M1-3) mark the characteristic frequency and its first harmonic, determined by the texture spatial period and the sliding speed.}
  \label{fig5}
\end{figure}

\begin{table}[!t]
  \caption{Spectral feature metrics of the three biomimetic fingertips and the human fingertip at the sliding speed of 20 mm/s. }
  \label{tab2}
  \begin{tabular*}{17.8cm}{@{\extracolsep{\fill}}l*{6}{>{\centering\arraybackslash}p{1.0cm}}*{3}{>{\centering\arraybackslash}p{1.9cm}}@{}}

    \hline
    & \multicolumn{6}{c}{Peak Prominence (-)} & \multicolumn{3}{c}{\makecell{150-300 Hz band energy\\($\times 10^3$ (mm/s$^2$)$^2$)}}
 \\
    \cline{2-7} \cline{8-10}
    & \multicolumn{2}{c}{M1: 0205 Sine} & \multicolumn{2}{c}{M2: 0210 Sine} & \multicolumn{2}{c}{M3: 0220 Sine} & M4: 005 Random & M5: 010 Random & M6: 020 Random \\
    Fingertip & $f_0$ & $2f_0$ & $f_0$ & $2f_0$ & $f_0$ & $2f_0$ & & & \\
    \hline
    PDMS & 1.98 & 2.76 & 2.46 & 2.53 & 4.21 & 1.66 & 5.37 & 6.24 & 4.08 \\
    TPU+Ecoflex & 5.05 & 2.80 & 2.50 & 6.00 & 3.10 & 3.44 & 484 & 436 & 474 \\
    PDMS+Ecoflex & 4.89 & 3.63 & 3.86 & 3.51 & 3.04 & 2.60 & 45.9 & 74.9 & 133 \\
    Human & 3.36 & 3.31 & 3.24 & 3.43 & 3.85 & 1.65 & 0.420 & 0.430 & 0.396 \\
    \hline
  \end{tabular*}
\end{table}

Spectral features of the vibrotactile signals offer insight into the influence of different fingertip designs on tactile transparency. Following the preprocessing described in Section 3.4, we computed the FFT amplitude spectrum and Welch's power spectral density (PSD) for each recording, as displayed in Figure 5.

As sliding speed increases, the overall vibration energy of the vibrotactile signals grows. Prior research has established that periodic textures generate characteristic spectral peaks at frequencies determined by the ratio of sliding speed to spatial period ($f = v/\lambda$) \cite{delhaye_2012}, and our observations are consistent with this: peaks shift upward with sliding speed and downward with spatial period, as marked by the dashed lines in Figure 5. As shown in Table 2, among the three biomimetic fingertips, the PDMS+Ecoflex design exhibits the highest average peak prominence at the characteristic frequency, with mean $f_0$ prominence of 3.93 across M1, M2, and M3, which produces spectra most similar to those of the human fingertip on periodic surfaces (Section 4.4). Peaks below 10 Hz are obscured by the DC side-lobes across all fingertips, and peak prominence generally becomes more pronounced as sliding speed increases.

The three random surfaces share a similar geometric configuration but differ in their RMS height distributions (0.05, 0.10, and 0.20 mm). For the PDMS+Ecoflex fingertip, vibration energy increases with surface RMS, particularly in the high-frequency range (150-300 Hz), with band energies of 45.9, 74.9, and 133.3 $\times 10^3$ (mm/s$^2$)$^2$ for M4, M5, and M6, respectively (Table 2). 
The PDMS fingertip produces spectral distributions more similar to those of the human fingertip: owing to the dominant low spatial frequency content of the random textures, both spectra exhibit prominent low-frequency components. The TPU+Ecoflex fingertip exhibits pronounced wideband vibrations, with 150-300 Hz band energy severalfold higher than that of the PDMS+Ecoflex design yet showing no systematic dependence on surface RMS (Table 2).

For the natural surfaces, Oxford cloth (M7) and leather (M8) share similar surface texture geometry, resulting in comparable spectral features across all four fingertips. Plant fiber (M9), characterized by its chaotic and rough texture, produces prominent high-frequency energy distributions for all fingertips except the PDMS. The two metal surfaces yield spectra dominated by DC side-lobes, with all three biomimetic fingertips exhibiting patterns similar to that of the human fingertip.

The spectral differences across the three biomimetic fingertips can be attributed to their distinct structural designs. The PDMS fingertip, lacking a multilayer mechanical gradient, provides limited amplification of texture-induced features with the lowest average $f_0$ prominence and band energy among the three designs (Table 2). The TPU+Ecoflex fingertip, with its stiffer TPU skin, produces rigid collision-like excitation when sliding over hard surfaces; combined with the material's low internal damping and the inherently low energy of characteristic-frequency peaks at low sliding speeds, the resulting wideband vibrations disproportionately mask the non-significant low-frequency peaks. The PDMS+Ecoflex fingertip, featuring a multilayer structure with a mechanical gradient, achieves wideband amplification of texture-induced vibrations, with pronounced peak distributions in the 150-300 Hz range that closely match those of the human fingertip.

\subsection{Spectral Similarity Analysis}
\begin{figure}[!t]
  \includegraphics[width=\linewidth]{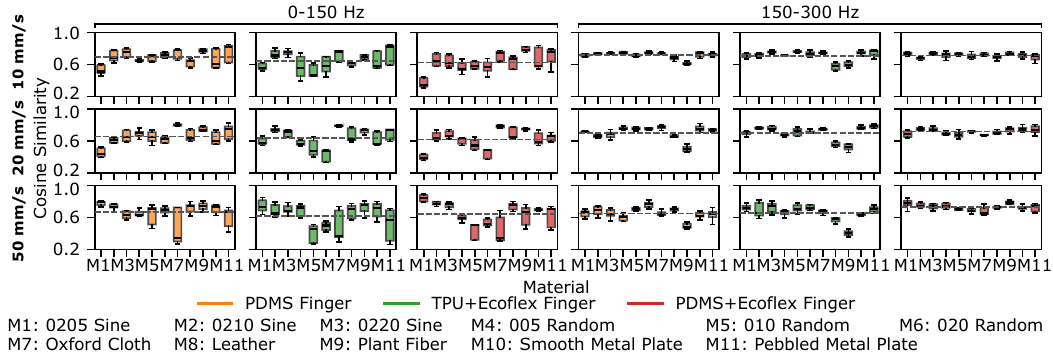}
  \caption{Spectral cosine similarity between each biomimetic fingertip and the human fingertip, computed on min-max normalized FFT spectra within two frequency bands (0-150 Hz and 150-300 Hz) across 11 textured surfaces at three sliding speeds (10, 20, and 50 mm/s); the grey dashed lines indicate the mean similarity across all 11 surfaces under each (finger, speed, band) condition. 
  }
  \label{fig6}
\end{figure}

To quantify the spectral similarity between biomimetic and human fingertips, we computed the cosine similarity between min-max normalized FFT spectra within two frequency bands (0--150 Hz and 150--300 Hz). For each condition (finger, surface, speed, and frequency band) , 9 similarity values were obtained from the 3 $\times$ 3 pairwise combinations of biomimetic and human repetitions, forming one box in Figure 6.

Overall, the three designs exhibit band-dependent similarity patterns (Figure 6, grey dashed lines). 
The PDMS fingertip attains the highest mean similarity scores in the 0--150 Hz band at all speeds: 0.69, 0.66, and 0.68 (10, 20, and 50 mm/s). In contrast, the PDMS+Ecoflex fingertip leads in the 150--300 Hz band at 20 and 50 mm/s, with similarity scores of 0.72 and 0.73, respectively. The advantage of PDMS+Ecoflex widens at 50 mm/s, with 0.73 versus 0.65 (PDMS) and 0.66 (TPU+Ecoflex). The category-level results are consistent with the spectral observations in Section 4.3. 
For the periodic surfaces (M1-3), the characteristic frequencies fall predominantly within the 0--150 Hz band; at 50 mm/s, these peaks become more pronounced and dominate the spectral energy distribution, leading to good similarity to the human fingertip for all three biomimetic designs: 0.80 (PDMS+Ecoflex), 0.72 (PDMS), and 0.69 (TPU+Ecoflex).


For plant fiber (M9), whose spectral features concentrate in the high-frequency range, the 150--300 Hz similarity analysis reveals that the PDMS+Ecoflex fingertip captures high-frequency characteristics more similar to the human fingertip than other robot fingertips, with median 0.75 versus 0.50 (PDMS) and 0.53 (TPU+Ecoflex) at 20 mm/s.

\section{User Study}
Ultimately, achieving high tactile transparency depends on whether the vibrotactile signals captured by the robot evoke a sensory experience similar to a direct human touch.
Sections 4.2 and 4.4 have analyzed the association between fingertip design and signal characteristics from both temporal and spectral domain perspectives, and compared them with vibrotactile signals acquired by the human fingertip. Yet, a comprehensive assessment of vibrotactile feedback must extend beyond purely physical characteristics to incorporate human perception, since structurally-distinct signals can still induce an authentic tactile experience by successfully aligning with a user's sensory expectations \cite{rosenkranz2023perceptual}.
Building on this, we conducted a user study to evaluate the tactile transparency of different fingertip designs from a perceptual perspective, establishing the association between fingertip design and perceptual characteristics.

\subsection{Experimental Setup}

\begin{figure}[!t]
  \includegraphics[width=\linewidth]{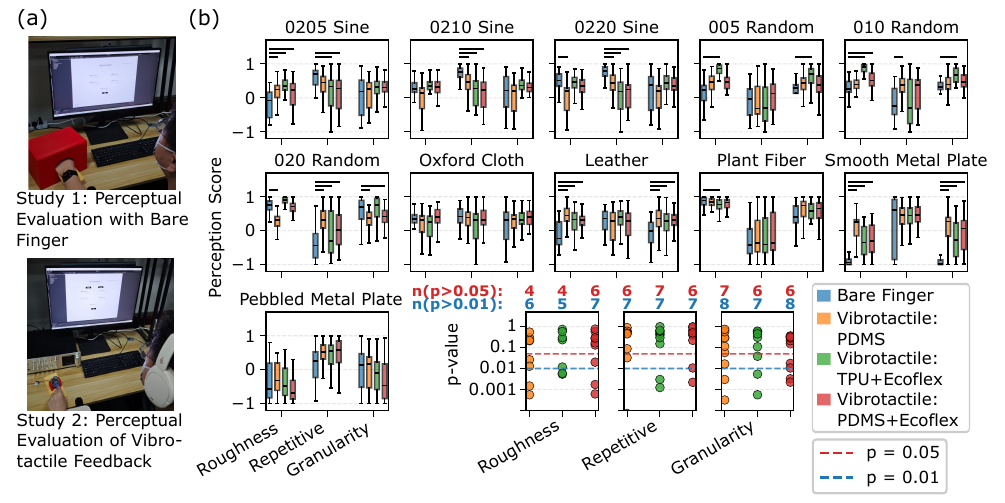}
  \caption{User study scenarios and perceptual evaluation. (a) Experimental setups for Test 1 (bare-hand exploration) and Test 2 (motor-delivered vibrotactile feedback). (b) Rating distributions from 30 participants across three biomimetic fingertips, 11 surfaces, and three perceptual dimensions (Roughness, Repetitive, Granularity). Scatter points indicate $p$-values from Wilcoxon signed-rank tests evaluating consistency with bare-hand ratings.
  }
  \label{fig7}
\end{figure}

To conduct the user study, we recruited a total of 30 participants (14 female, 16 male); all were right-handed and had very limited experience with haptic devices. All participants reported no physical or cognitive impairments that might interfere with the results, and gave their written informed consent prior to the experiment. 
The experimental protocol was approved by the Institutional Review Board of Harbin Institute of Technology (No. HIT-2024046), and all participants received financial compensation. 

The user study was implemented through a custom Unity-based interactive interface. Vibrotactile feedback was delivered by a voice coil motor (4$\Omega$25W, Qianyun) mounted horizontally via a custom 3D-printed fixture, with its vibration axis aligned along the participant's finger. Participants wore noise-canceling headphone (W860NB Pro, EDIFIER) throughout the experiment to eliminate auditory cues. 
Perceptual ratings were collected across three dimensions—Roughness, Repetitiveness, and Granularity—using a continuous scale from -1 to 1. Prior to the experiment, participants were informed that -1 represented no perception at all, while 1 represented the strongest sensation imaginable.

At the beginning of the experiment, participants were first briefed on the protocol and completed a practice session. The experiment consisted of two sequential tests, as illustrated in Figure 7(a). In Test 1, participants directly explored the 11 textured surfaces using their left index finger. To ensure total visual deprivation, the surfaces were placed inside a light-tight chamber, and participants performed back-and-forth strokes along the longitudinal axis of the finger. Each surface was presented twice in a randomized order, resulting in a total of 22 trials per participant.
The evaluation result of Test 1 is shown as the blue boxes in Figure 7(b). In Test 2, participants received vibrotactile feedback via the motor, which rendered acceleration signals previously acquired from three biomimetic fingertips sliding across the 11 surfaces at 20 mm/s. Each of the 33 finger-surface combinations was presented twice in random order, yielding 66 trials per participant.

To evaluate the tactile transparency between biomimetic feedback and direct bare-hand contact, paired samples were constructed by matching each participant's Test 1 (bare-hand) rating with their corresponding Test 2 (vibrotactile feedback) rating across all combinations of fingertips, surfaces, and dimensions. A Wilcoxon signed-rank test was conducted on these paired samples, yielding 99 $p$-values (3 fingertips $\times$ 11 surfaces $\times$ 3 dimensions). 
Note that, in this study, higher $p$-values indicate greater perceptual consistency, as they reflect a lack of statistically significant differences in perceptual ratings between the two modalities.
Figure 7(b) depicts the rating distributions alongside the resulting $p$-value distribution.

\subsection{Evaluation of Tactile Transparency}

The perceptual ratings demonstrate distinct patterns across surface categories and fingertip designs (Figure 7(b)). On periodic surfaces (M1-3), vibrotactile feedback from all three biomimetic fingertips conveys a perceptible sense of periodicity in the Repetitive dimension, consistently weaker than that perceived through bare-hand exploration; among the three designs, the PDMS fingertip yields Repetitive ratings that align most closely with bare-hand perception. On random surfaces (M4-6), the three designs differ markedly: the Roughness ratings of the PDMS+Ecoflex fingertip increase with surface RMS height; the PDMS fingertip shows minimal variation in Roughness across the three surfaces, yet delivers Granularity ratings consistent with bare-hand perception; in contrast, the TPU+Ecoflex fingertip elicits pronounced Roughness and Granularity sensations across all three surfaces.

On natural surfaces (M7-11), the vibrotactile feedback generally shows strong agreement with bare-hand perception. 
For Oxford cloth (M7) and leather (M8), all three biomimetic fingertips yield Repetitive ratings that closely mirror those of the bare hand, with no statistically significant differences between them ($p > 0.05$).
On plant fiber (M9), the vibrotactile feedback is broadly consistent with direct exploration: all three fingertips match bare-hand ratings at p $>$ 0.01 across the three dimensions, with the sole exception of the TPU+Ecoflex fingertip in Roughness ($p = 0.006$). The smooth metal plate (M10) highlights a key limitation: the biomimetic feedback differs from bare-hand perception in both Roughness and Granularity ($p < 0.001$ for all three fingers), revealing compromised tactile transparency on near-featureless surfaces. 
By contrast, on the pebbled metal plate (M11), Roughness ratings are comparable to the bare hand for all three fingertips ($p > 0.05$), though the PDMS fingertip shows a marginal difference in Repetitive ($p = 0.033$) and the PDMS+Ecoflex fingertip in Granularity ($p = 0.011$).

The scatter points in Figure 7(b) provide a summary of these comparisons: for each fingertip, the number of surface-dimension combinations with $p > 0.05$ serves as an overall measure of perceptual consistency with bare-hand ratings. 
The PDMS+Ecoflex fingertip leads in Roughness (6/11 surfaces with $p > 0.05$), the TPU+Ecoflex fingertip in Repetitive (7/11), and the PDMS fingertip in Granularity (7/11). 
The Repetitive advantage of the TPU+Ecoflex fingertip arises mainly from the non-periodic surfaces (7/8 match). 
On periodic surfaces, feedback from the PDMS fingertip is the closest to bare-hand perception.

Taken together, these dimension-specific advantages reveal complementary perceptual profiles: the PDMS+Ecoflex fingertip tracks surface roughness through its ratings, the TPU+Ecoflex fingertip over-renders roughness and granularity on random surfaces, and the PDMS fingertip maintains stable granularity and periodicity perception across surface categories. On periodic surfaces, the perceived periodicity is systematically attenuated compared to bare-hand touch, even though the recorded signals retain characteristic spectral peaks (Section 4.3). On the smooth metal plate, the vibrotactile feedback evokes texture sensations that are absent during direct skin contact. These findings are discussed in Section.~\ref{sec:discussion} through a cross-level synthesis connecting signal characteristics (Sections 4.2 and 4.3) to the present perceptual outcomes.

\section{Discussion}\label{sec:discussion}

The temporal, spectral, and perceptual analysis results of this study establish a transmission chain from fingertip design to signal characteristics and, ultimately, to perceived tactile transparency. This chain not only reveals how individual design features shape tactile transparency but also provides a basis for designing biomimetic fingertips toward this target. Building on the results of Sections 4 and 5.2, this section attributes the perceptual differences among the three designs to specific design features and traces the anomalies shared across designs to factors at the system level.

Among the design features examined, the multilayer mechanical gradient yields the clearest improvements in signal intensity and texture-feature representation. Relative to the single-material PDMS design, the PDMS+Ecoflex fingertip generates more intense vibrations (Section 4), exhibits higher sensitivity to characteristic spectral peaks, and produces signal energy that varies with surface roughness (Table 2). Perceptually, the PDMS+Ecoflex fingertip reproduces the bare-hand trend of increasing Roughness ratings with surface RMS height (Section 5.2), consistent with the signal-level variation trend, as the gradient design strengthens texture-related spectral features. In contrast, the single-material PDMS design provides limited amplification, with the lowest band energy among the three designs (Table 2). Notably, the intense vibrations of the TPU+Ecoflex fingertip arise not only from the same multilayer gradient but also from the high Young's modulus of its skin layer. 

Although the stiff TPU skin amplifies vibration intensity, the recorded signals fail to accurately capture the features necessary for realistic texture rendering.
At the signal level, the characteristic spectral peaks of the TPU+Ecoflex fingertip are masked by the elevated wideband energy (Table 2). Perceptually, this design elicits consistently high Roughness and Granularity ratings across all random surfaces with poor discrimination (Section 5.2). Furthermore, its advantage in the Repetitive dimension stems primarily from non-periodic surfaces (7/8 match; Section 5.2). These phenomena stem from the high stiffness of the TPU skin, which induces rigid collision-like excitation when sliding over hard surfaces; coupled with the material's low internal damping, these intense wideband vibrations propagate to the sensor with minimal attenuation.

Beyond the design-specific effects discussed above, the results indicate that signal similarity does not fully determine tactile transparency. Sections 4 and 5 show that while the PDMS and human fingertip signals are highly similar, the feedback cannot convey roughness information comparable to bare-hand perception, possibly because neither the PDMS fingertip nor a human-fingerpad accelerometer captures vibration features related to surface roughness changes. Section 4 shows that all three fingertips capture material-related characteristic spectral peaks (Table 2), yet Section 5 reveals that periodicity perception is consistently weaker than bare-hand perception across the three designs, presumably because harmonics blur the periodic features and, in the TPU+Ecoflex fingertip, the peaks are masked by wideband energy. 
On the smooth metal plate (M10), the acquired spectra introduce signal features unrelated to the smooth surface, which evoke texture sensations absent in direct contact (Section 5.2) despite the spectra being similar to that of the human fingertip (Section 4.3), with plausible origins including residual side-lobes produced by robot movement that persist even after DC removal filtering, which are raised into the perceptible range by the group-wise RMS alignment. 

Together, these cases indicate that tactile transparency is a dimension-specific read-out of the signal: the Roughness dimension draws on the variation trend of signal energy with surface roughness, the Granularity dimension on the overall spectral shape, and the Repetitive dimension on the faithful representation of characteristic peaks.

The observations above can be traced to three constraints of the present system. First, the single-contact-point feedback cannot convey the spatially distributed mechanoreceptive cues that contribute to periodicity perception in direct touch; future work could employ multi-actuator arrays to restore spatial information. Second, the feedback was limited to a single perceptual modality of vibration acceleration, lacking the force, temperature, and skin-stretch cues that the bare hand integrates for texture perception; future work could combine multi-modal feedback to enrich the perceptual space. Third, the high-sensitivity IMU captures system noise such as DC side-lobes, which introduce non-texture-related signals into the acquired data; future work could improve signal conditioning and noise isolation to reduce these artifacts. 

\section{Conclusion}
This paper presented a biomimetic fingertip that replicates the multilayer mechanical gradient and fingerprint morphology of the human finger, embedding a high-sensitivity IMU to capture texture-induced vibrations for remote vibrotactile rendering. Three variants were systematically compared with the human fingertip through temporal and spectral analyses and a user study spanning three perceptual dimensions, establishing an information transmission chain spanning fingertip design, signal characteristics, and, ultimately, perceived tactile feedback.
Within this chain, the multilayer mechanical gradient yields the clearest improvements in signal intensity and texture-feature representation, whereas a stiffer skin layer amplifies vibration intensity at the expense of feature representation. As the perceptual dimensions draw on distinct signal attributes: roughness on energy scaling, granularity on the overall spectral shape, and repetitivity on the faithful representation of characteristic peaks; designs should therefore be matched to the dimensions prioritized by the application. The remaining gaps lie in the acquisition and rendering stages rather than the fingertip designs, pointing to future work on multi-actuator spatial rendering, cross-speed perceptual evaluation, and data-driven rendering approach for improved tactile-transparent teleoperation.

\medskip
\textbf{Supporting Information} \par 
Supporting Information is available from the Wiley Online Library or from the author.

\medskip
\textbf{Acknowledgements} \par 
This work was supported by National Natural Science Foundation of China under Grant 62576119 and Shenzhen Science and Technology Innovation Program under Grant JCYJ20241202123716021.
\medskip

\bibliographystyle{MSP}
\bibliography{v8_references}

@article{Andrussow_2023,
  title = {Minsight: A Fingertip‐Sized Vision‐Based Tactile Sensor for Robotic Manipulation},
  volume = {5},
  ISSN = {2640-4567},
  url = {http://dx.doi.org/10.1002/aisy.202300042},
  DOI = {10.1002/aisy.202300042},
  number = {8},
  journal = {Advanced Intelligent Systems},
  publisher = {Wiley},
  author = {Andrussow, Iris and Sun, Huanbo and Kuchenbecker, Katherine J. and Martius, Georg},
  year = {2023},
}

@article{Bai_2023,
  title = {A robotic sensory system with high spatiotemporal resolution for texture recognition},
  volume = {14},
  ISSN = {2041-1723},
  url = {http://dx.doi.org/10.1038/s41467-023-42722-4},
  DOI = {10.1038/s41467-023-42722-4},
  number = {1},
  journal = {Nature Communications},
  publisher = {Springer Science and Business Media LLC},
  author = {Bai, Ningning and Xue, Yiheng and Chen, Shuiqing and Shi, Lin and Shi, Junli and Zhang, Yuan and Hou, Xingyu and Cheng, Yu and Huang, Kaixi and Wang, Weidong and Zhang, Jin and Liu, Yuan and Guo, Chuan Fei},
  year = {2023},
}

@article{Bernard_2024,
  title = {The High/Low Frequency Balance Drives Tactile Perception of Noisy Vibrations},
  volume = {17},
  ISSN = {2334-0134},
  url = {http://dx.doi.org/10.1109/TOH.2024.3371264},
  DOI = {10.1109/toh.2024.3371264},
  number = {4},
  journal = {IEEE Transactions on Haptics},
  publisher = {Institute of Electrical and Electronics Engineers (IEEE)},
  author = {Bernard, Corentin and Thoret, Etienne and Huloux, Nicolas and Ystad, Sølvi},
  year = {2024},
  pages = {614–624},
}

@article{Cei_2025,
  title = {A review on finite element modelling of finger and hand mechanical behaviour in haptic interactions},
  volume = {24},
  ISSN = {1617-7940},
  url = {http://dx.doi.org/10.1007/s10237-025-01943-w},
  DOI = {10.1007/s10237-025-01943-w},
  number = {3},
  journal = {Biomechanics and Modeling in Mechanobiology},
  publisher = {Springer Science and Business Media LLC},
  author = {Cei, Gianmarco and Artoni, Alessio and Bianchi, Matteo},
  year = {2025},
  pages = {895–917},
}

@article{Choi_2019,
  title = {Biomimetic Tactile Sensors with Bilayer Fingerprint Ridges Demonstrating Texture Recognition},
  volume = {10},
  ISSN = {2072-666X},
  url = {http://dx.doi.org/10.3390/mi10100642},
  DOI = {10.3390/mi10100642},
  number = {10},
  journal = {Micromachines},
  publisher = {MDPI AG},
  author = {Choi, Eunsuk and Sul, Onejae and Lee, Jusin and Seo, Hojun and Kim, Sunjin and Yeom, Seongoh and Ryu, Gunwoo and Yang, Heewon and Shin, Yoonsoo and Lee, Seung-Beck},
  year = {2019},
  pages = {642},
}

@INPROCEEDINGS{Dai_2022,
  author={Dai, Kevin and Wang, Xinyu and Rojas, Allison M. and Harber, Evan and Tian, Yu and Paiva, Nicholas and Gnehm, Joseph and Schindewolf, Evan and Choset, Howie and Webster-Wood, Victoria A. and Li, Lu},
  booktitle={2022 International Conference on Robotics and Automation (ICRA)}, 
  title={Design of a Biomimetic Tactile Sensor for Material Classification}, 
  year={2022},
  volume={},
  number={},
  pages={10774-10780},
  doi={10.1109/ICRA46639.2022.9811543}}

@article{Fagiani_2012,
  title = {Contact of a Finger on Rigid Surfaces and Textiles: Friction Coefficient and Induced Vibrations},
  volume = {48},
  ISSN = {1573-2711},
  url = {http://dx.doi.org/10.1007/s11249-012-0010-0},
  DOI = {10.1007/s11249-012-0010-0},
  number = {2},
  journal = {Tribology Letters},
  publisher = {Springer Science and Business Media LLC},
  author = {Fagiani, Ramona and Massi, Francesco and Chatelet, Eric and Costes, Jean Philippe and Berthier, Yves},
  year = {2012},
  pages = {145–158},
}

@article{Felicetti_2023,
  title = {Tactile discrimination of real and simulated isotropic textures by Friction-Induced Vibrations},
  volume = {184},
  ISSN = {0301-679X},
  url = {http://dx.doi.org/10.1016/j.triboint.2023.108443},
  DOI = {10.1016/j.triboint.2023.108443},
  journal = {Tribology International},
  publisher = {Elsevier BV},
  author = {Felicetti, Livia and Sutter, Chloé and Chatelet, Eric and Latour, Antoine and Mouchnino, Laurence and Massi, Francesco},
  year = {2023},
  pages = {108443},
}

@article{Geng_2023,
  title = {Construction of Wearable Touch Sensors by Mimicking the Properties of Materials and Structures in Nature},
  volume = {8},
  ISSN = {2313-7673},
  url = {http://dx.doi.org/10.3390/biomimetics8040372},
  DOI = {10.3390/biomimetics8040372},
  number = {4},
  journal = {Biomimetics},
  publisher = {MDPI AG},
  author = {Geng, Baojun and Zeng, Henglin and Luo, Hua and Wu, Xiaodong},
  year = {2023},
  pages = {372},
}

@article{Guo_2024,
  title = {Zero‐Biased Bionic Fingertip E‐Skin with Multimodal Tactile Perception and Artificial Intelligence for Augmented Touch Awareness},
  volume = {36},
  ISSN = {1521-4095},
  url = {http://dx.doi.org/10.1002/adma.202406778},
  DOI = {10.1002/adma.202406778},
  number = {39},
  journal = {Advanced Materials},
  publisher = {Wiley},
  author = {Guo, Xinge and Sun, Zhongda and Zhu, Yao and Lee, Chengkuo},
  year = {2024},
}

@article{Hao_2022,
  title = {Fingerprint-inspired surface texture for the enhanced tip pinch performance of a soft robotic hand in lubricated conditions},
  volume = {11},
  ISSN = {2223-7704},
  url = {http://dx.doi.org/10.1007/s40544-022-0688-4},
  DOI = {10.1007/s40544-022-0688-4},
  number = {7},
  journal = {Friction},
  publisher = {Tsinghua University Press},
  author = {Hao, Tianze and Xiao, Huaping and Liu, Shuhai and Liu, Yibo},
  year = {2022},
  pages = {1349–1358},
}

@article{Hao_2024,
  title = {Friction Enhancement Through Fingerprint-like Soft Surface Textures in Soft Robotic Grippers for Grasping Abilities},
  volume = {72},
  ISSN = {1573-2711},
  url = {http://dx.doi.org/10.1007/s11249-024-01848-2},
  DOI = {10.1007/s11249-024-01848-2},
  number = {2},
  journal = {Tribology Letters},
  publisher = {Springer Science and Business Media LLC},
  author = {Hao, Tianze and Xiao, Huaping and Wang, Jutao and Wang, Xiaofei and Liu, Shuhai and Liu, Qingjian},
  year = {2024},
}

@article{Hao_2025,
  title = {3D printed hydrogel with fingerprint-inspired features for enhanced sensing performance in electronic skin},
  volume = {525},
  ISSN = {1385-8947},
  url = {http://dx.doi.org/10.1016/j.cej.2025.170534},
  DOI = {10.1016/j.cej.2025.170534},
  journal = {Chemical Engineering Journal},
  publisher = {Elsevier BV},
  author = {Hao, Jianshe and Xu, Yanbin and Yang, Chaofan and Guo, Rui and Lyu, Yang and Ji, Zhongying and Wang, Xiaolong},
  year = {2025},
  pages = {170534},
}

@article{Hou_2024,
  title = {Biometric‐Tuned E‐Skin Sensor with Real Fingerprints Provides Insights on Tactile Perception: Rosa Parks Had Better Surface Vibrational Sensation than Richard Nixon},
  volume = {11},
  ISSN = {2198-3844},
  url = {http://dx.doi.org/10.1002/advs.202400234},
  DOI = {10.1002/advs.202400234},
  number = {34},
  journal = {Advanced Science},
  publisher = {Wiley},
  author = {Hou, Senlin and Huang, Qingyun and Zhang, Hongyu and Chen, Qingjiu and Wu, Cong and Wu, Mengge and Meng, Chen and Yao, Kuanming and Yu, Xinge and Roy, Vellaisamy A. L. and Daoud, Walid and Wang, Jianping and Li, Wen Jung},
  year = {2024},
}

@article{Jarocka_2021,
  title = {Human Touch Receptors Are Sensitive to Spatial Details on the Scale of Single Fingerprint Ridges},
  volume = {41},
  ISSN = {1529-2401},
  url = {http://dx.doi.org/10.1523/JNEUROSCI.1716-20.2021},
  DOI = {10.1523/jneurosci.1716-20.2021},
  number = {16},
  journal = {The Journal of Neuroscience},
  publisher = {Society for Neuroscience},
  author = {Jarocka, Ewa and Pruszynski, J. Andrew and Johansson, Roland S.},
  year = {2021},
  pages = {3622–3634},
}

@article{Kang_2024,
  title = {Biomimic and bioinspired soft neuromorphic tactile sensory system},
  volume = {11},
  ISSN = {1931-9401},
  url = {http://dx.doi.org/10.1063/5.0204104},
  DOI = {10.1063/5.0204104},
  number = {2},
  journal = {Applied Physics Reviews},
  publisher = {AIP Publishing},
  author = {Kang, Kyowon and Kim, Kiho and Baek, Junhyeong and Lee, Doohyun J. and Yu, Ki Jun},
  year = {2024},
}

@article{Kim_2020,
  title = {Biomimetic Hybrid Tactile Sensor with Ridged Structure That Mimics Human Fingerprints to Acquire Surface Texture Information},
  volume = {32},
  ISSN = {0914-4935},
  url = {http://dx.doi.org/10.18494/SAM.2020.2995},
  DOI = {10.18494/sam.2020.2995},
  number = {11},
  journal = {Sensors and Materials},
  publisher = {MYU K.K.},
  author = {Kim, Sung Joon and Choi, Jae Young and Moon, Hyungpil and Choi, Hyouk Ryeol and Koo, Ja Choon},
  year = {2020},
  pages = {3787},
}

@INPROCEEDINGS{Lee_2024,
  author={Lee, Hyung-Tak and Bak, Keungyonh and Chun, Sungwoo and Hwang, Han-Jeong},
  booktitle={2024 International Conference on Cyberworlds (CW)}, 
  title={Classification of Human Tactile Perception Using Vibration Data}, 
  year={2024},
  volume={},
  number={},
  pages={183-186},
  doi={10.1109/CW64301.2024.00054}}

@article{Navaraj_2019,
  title = {Fingerprint‐Enhanced Capacitive‐Piezoelectric Flexible Sensing Skin to Discriminate Static and Dynamic Tactile Stimuli},
  volume = {1},
  ISSN = {2640-4567},
  url = {http://dx.doi.org/10.1002/aisy.201900051},
  DOI = {10.1002/aisy.201900051},
  number = {7},
  journal = {Advanced Intelligent Systems},
  publisher = {Wiley},
  author = {Navaraj, William and Dahiya, Ravinder},
  year = {2019},
}

@article{Ouyang_2024,
  title = {Artificial Tactile Sensory Finger for Contact Pattern Identification Based on High Spatiotemporal Piezoresistive Sensor Array},
  volume = {16},
  ISSN = {1944-8252},
  url = {http://dx.doi.org/10.1021/acsami.4c07056},
  DOI = {10.1021/acsami.4c07056},
  number = {44},
  journal = {ACS Applied Materials Interfaces},
  publisher = {American Chemical Society (ACS)},
  author = {Ouyang, Qiangqiang and Wang, Xiaoying and Wang, Shaoyi and Huang, Zizhen and Shi, Zhaohui and Pang, Mao and Liu, Bin and Tan, Chee Keong and Yang, Qintai and Rong, Limin},
  year = {2024},
  pages = {61179–61193},
}

@article{Qin_2023,
  title = {Perception of Static and Dynamic Forces with a Bio-inspired Tactile Fingertip},
  volume = {20},
  ISSN = {2543-2141},
  url = {http://dx.doi.org/10.1007/s42235-023-00344-y},
  DOI = {10.1007/s42235-023-00344-y},
  number = {4},
  journal = {Journal of Bionic Engineering},
  publisher = {Springer Science and Business Media LLC},
  author = {Qin, Longhui and Shi, Xiaowei and Wang, Yihua and Zhou, Zhitong},
  year = {2023},
  pages = {1544–1554},
}

@article{Qin_2024,
  title = {Fingerprint-inspired biomimetic tactile sensors for the surface texture recognition},
  volume = {371},
  ISSN = {0924-4247},
  url = {http://dx.doi.org/10.1016/j.sna.2024.115275},
  DOI = {10.1016/j.sna.2024.115275},
  journal = {Sensors and Actuators A: Physical},
  publisher = {Elsevier BV},
  author = {Qin, Liguo and Hao, Luxin and Huang, Xiaodong and Zhang, Rui and Lu, Shan and Wang, Zheng and Liu, Jianbo and Ma, Zeyu and Xia, Xiaohua and Dong, Guangneng},
  year = {2024},
  pages = {115275},
}

@article{Rostamian_2022,
  title = {Texture recognition based on multi-sensory integration of proprioceptive and tactile signals},
  volume = {12},
  ISSN = {2045-2322},
  url = {http://dx.doi.org/10.1038/s41598-022-24640-5},
  DOI = {10.1038/s41598-022-24640-5},
  number = {1},
  journal = {Scientific Reports},
  publisher = {Springer Science and Business Media LLC},
  author = {Rostamian, Behnam and Koolani, MohammadReza and Abdollahzade, Pouya and Lankarany, Milad and Falotico, Egidio and Amiri, Mahmood and V. Thakor, Nitish},
  year = {2022},
}

@article{Serhat_2021,
  title = {Free and Forced Vibration Modes of the Human Fingertip},
  volume = {11},
  ISSN = {2076-3417},
  url = {http://dx.doi.org/10.3390/app11125709},
  DOI = {10.3390/app11125709},
  number = {12},
  journal = {Applied Sciences},
  publisher = {MDPI AG},
  author = {Serhat, Gokhan and Kuchenbecker, Katherine J.},
  year = {2021},
  pages = {5709},
}

@article{Serhat_2024,
  title = {Fingertip dynamic response simulated across excitation points and frequencies},
  volume = {23},
  ISSN = {1617-7940},
  url = {http://dx.doi.org/10.1007/s10237-024-01844-4},
  DOI = {10.1007/s10237-024-01844-4},
  number = {4},
  journal = {Biomechanics and Modeling in Mechanobiology},
  publisher = {Springer Science and Business Media LLC},
  author = {Serhat, Gokhan and Kuchenbecker, Katherine J.},
  year = {2024},
  pages = {1369–1376},
}

@INPROCEEDINGS{Tanaka_2019,
  author={Tanaka, Yoshihiro and Hasegawa, Tatsuya and Hashimoto, Masatoshi and Igarashi, Takanori},
  booktitle={2019 IEEE World Haptics Conference (WHC)}, 
  title={Artificial Fingers Wearing Skin Vibration Sensor for Evaluating Tactile Sensations}, 
  year={2019},
  volume={},
  number={},
  pages={377-382},
  doi={10.1109/WHC.2019.8816155}}

@article{Weiland_2024,
  title = {Tactile simulation of textile fabrics: Design of simulation signals with regard to fingerprint},
  volume = {191},
  ISSN = {0301-679X},
  url = {http://dx.doi.org/10.1016/j.triboint.2023.109113},
  DOI = {10.1016/j.triboint.2023.109113},
  journal = {Tribology International},
  publisher = {Elsevier BV},
  author = {Weiland, Benjamin and Leclinche, Floriane and Kaci, Anis and Camillieri, Brigitte and Lemaire-Semail, Betty and Bueno, Marie-Ange},
  year = {2024},
  pages = {109113},
}

@article{Wu_2018,
  title = {A skin-inspired tactile sensor for smart prosthetics},
  volume = {3},
  ISSN = {2470-9476},
  url = {http://dx.doi.org/10.1126/scirobotics.aat0429},
  DOI = {10.1126/scirobotics.aat0429},
  number = {22},
  journal = {Science Robotics},
  publisher = {American Association for the Advancement of Science (AAAS)},
  author = {Wu, Yuanzhao and Liu, Yiwei and Zhou, Youlin and Man, Qikui and Hu, Chao and Asghar, Waqas and Li, Fali and Yu, Zhe and Shang, Jie and Liu, Gang and Liao, Meiyong and Li, Run-Wei},
  year = {2018},
}

@article{Xu_2026,
  title = {A Fingertip-Size Six-Axis Force Sensor via Origami Coil Arrays for Intrinsic Tactile Sensing},
  ISSN = {1941-014X},
  url = {http://dx.doi.org/10.1109/TMECH.2026.3660186},
  DOI = {10.1109/tmech.2026.3660186},
  journal = {IEEE/ASME Transactions on Mechatronics},
  publisher = {Institute of Electrical and Electronics Engineers (IEEE)},
  author = {Xu, Yingao and Wu, Houping and Xie, Yunfei and Zhang, Jiayuan and Wang, Hongbo},
  year = {2026},
  pages = {1–11},
}

@article{Zhang_2022,
  title = {Finger-inspired rigid-soft hybrid tactile sensor with superior sensitivity at high frequency},
  volume = {13},
  ISSN = {2041-1723},
  url = {http://dx.doi.org/10.1038/s41467-022-32827-7},
  DOI = {10.1038/s41467-022-32827-7},
  number = {1},
  journal = {Nature Communications},
  publisher = {Springer Science and Business Media LLC},
  author = {Zhang, Jinhui and Yao, Haimin and Mo, Jiaying and Chen, Songyue and Xie, Yu and Ma, Shenglin and Chen, Rui and Luo, Tao and Ling, Weisong and Qin, Lifeng and Wang, Zuankai and Zhou, Wei},
  year = {2022},
}

@article{Zhao_2021,
  title = {Fingerprint-inspired electronic skin based on triboelectric nanogenerator for fine texture recognition},
  volume = {85},
  ISSN = {2211-2855},
  url = {http://dx.doi.org/10.1016/j.nanoen.2021.106001},
  DOI = {10.1016/j.nanoen.2021.106001},
  journal = {Nano Energy},
  publisher = {Elsevier BV},
  author = {Zhao, Xuan and Zhang, Zheng and Xu, Liangxu and Gao, Fangfang and Zhao, Bin and Ouyang, Tian and Kang, Zhuo and Liao, Qingliang and Zhang, Yue},
  year = {2021},
  pages = {106001},
}

@article{johansson_2009,
  title={Coding and use of tactile signals from the fingertips in object manipulation tasks},
  author={Johansson, Roland S and Flanagan, J Randall},
  journal={Nature Reviews Neuroscience},
  volume={10},
  number={5},
  pages={345--359},
  year={2009},
  publisher={Nature Publishing Group},
  doi={10.1038/nrn2621}
}

@article{Shi_2023, 
     title={Surface Recognition With a Bioinspired Tactile Fingertip}, 
     volume={23}, 
     ISSN={2379-9153}, 
     url={http://dx.doi.org/10.1109/JSEN.2023.3291720}, 
     DOI={10.1109/jsen.2023.3291720}, 
     number={16}, 
     journal={IEEE Sensors Journal}, 
     publisher={Institute of Electrical and Electronics Engineers (IEEE)}, 
     author={Shi, Xiaowei and Wang, Yihua and Qin, Longhui}, 
     year={2023}, 
     month=Aug, 
     pages={18842–18855} 
 }

@article{Qiu_2020, 
  title={Bioinspired, multifunctional dual-mode pressure sensors as electronic skin for decoding complex loading processes and human motions},
  author={Qiu, Ye and Tian, Ye and Sun, Shenshen and Hu, Jiahui and Wang, Youyan and Zhang, Zheng and Liu, Aiping and Cheng, Huanyu and Gao, Weizhan and Zhang, Wenan and others},
  journal={Nano Energy},
  volume={78},
  pages={105337},
  year={2020},
  publisher={Elsevier}
}

@article{sankar_2025,
  title={A natural biomimetic prosthetic hand with neuromorphic tactile sensing for precise and compliant grasping},
  author={Sankar, Sriramana and Cheng, Wen-Yu and Zhang, Jinghua and Slepyan, Ariel and Iskarous, Mark M and Greene, Rebecca J and DeBrabander, Rene and Chen, Junjun and Gupta, Arnav and Thakor, Nitish V},
  journal={Science Advances},
  volume={11},
  number={10},
  pages={eadr9300},
  year={2025},
  publisher={American Association for the Advancement of Science}
}

@article{Qiu_2024_sciadv,
  title={Quantitative softness and texture bimodal haptic sensors for robotic clinical feature identification and intelligent picking},
  author={Qiu, Ye and Wang, Fangnan and Zhang, Zhuang and Shi, Kuanqiang and Song, Yi and Lu, Jiutian and Xu, Minjia and Qian, Mengyuan and Zhang, Wenan and Wu, Jixuan and others},
  journal={Science Advances},
  volume={10},
  number={30},
  pages={eadp0348},
  year={2024},
  publisher={American Association for the Advancement of Science}
}

@article{kovenburg_2023,
  title={The orientation dependence of the fingerprint effect for slip speed estimation and control},
  author={Kovenburg, Robert and Slezak, Andrew and George, Chase and Gale, Richard and Aksak, Burak},
  journal={IEEE Sensors Journal},
  volume={23},
  number={5},
  pages={5437--5447},
  year={2023},
  publisher={IEEE}
}

@article{somer_2015,
  title={A multi-scale computational assessment of channel gating assumptions within the Meissner corpuscle},
  author={Somer, DD and Peri{\'c}, D and de Souza Neto, EA and Dettmer, WG},
  journal={Journal of Biomechanics},
  volume={48},
  number={1},
  pages={73--80},
  year={2015},
  publisher={Elsevier}
}

@article{delhaye_2012,
  title={Texture-induced vibrations in the forearm during tactile exploration},
  author={Delhaye, Benoit and Hayward, Vincent and Lef{\`e}vre, Philippe and Thonnard, Jean-Louis},
  journal={Frontiers in behavioral neuroscience},
  volume={6},
  pages={37},
  year={2012},
  publisher={Frontiers Media SA}
}

@article{Qiao_2023,
  title={Non-equilibrium-growing aesthetic ionic skin for fingertip-like strain-undisturbed tactile sensation and texture recognition},
  author={Qiao, Haiyan and Sun, Shengtong and Wu, Peiyi},
  journal={Advanced materials},
  volume={35},
  number={21},
  pages={2300593},
  year={2023},
  publisher={Wiley Online Library}
}

@article{rosenkranz2023perceptual,
  title={A perceptual model-based approach to plausible authoring of vibration for the haptic metaverse},
  author={Rosenkranz, Robert and Altinsoy, M Ercan},
  journal={IEEE Transactions on Haptics},
  volume={17},
  number={2},
  pages={263--276},
  year={2023},
  publisher={IEEE}
}

\end{document}